\documentclass[lettersize, conference]{IEEEtran}
\IEEEoverridecommandlockouts

\usepackage[letterpaper]{geometry}

\usepackage{graphicx}
\usepackage{svg}
\usepackage{subfig}
\graphicspath{{figs/}}

\usepackage{multirow}
\usepackage{multicol}
\usepackage{amsmath,amssymb,amsfonts}
\usepackage{algorithm2e}
\usepackage{gensymb}
\usepackage{xcolor}
\usepackage[colorlinks=true, linkcolor=black, 
            urlcolor=cyan, filecolor=black,
            citecolor=black]{hyperref}
\usepackage{url}
\usepackage{amsmath,bm}
\usepackage{comment}
\usepackage{algorithm2e}
\usepackage{float} 
\usepackage{cite}
\usepackage{mathptmx} 

    \title{\vspace{10pt}Soft Finger Grasp Force and Contact State Estimation from Tactile Sensors}
\author{Hun Jang$^{1}$, Joonbum Bae$^{2*}$, Kevin Haninger$^{1*}$ 
\thanks{*This work was supported by European Union's Horizon 2020 research and innovation programme under grant agreement No. 101058521 — CONVERGING.}
\thanks{$^{1}$Department of Automation, Fraunhofer IPK, Berlin, Germany
        {\tt\small firstname.lastname@ipk.fraunhofer.de}}%
\thanks{$^{2}$ School of Mechanical Engineering, Korea University, Seoul, Korea {\tt\small jbbae@korea.ac.kr}}
\thanks{$^{*}$ denotes equal contribution as co-corresponding authors.}}
\begin{document}
\maketitle 
\begin{abstract}
Soft robotic fingers can improve adaptability in grasping and manipulation, compensating for geometric variation in object or environmental contact, but today lack force capacity and fine dexterity. Integrated tactile sensors can provide grasp and task information which can improve dexterity, but should ideally not require object-specific training. The total force vector exerted by a finger provides general information to the internal grasp forces (e.g. for grasp stability) and, when summed over fingers, an estimate of the external force acting on the grasped object (e.g. for task-level control). 
In this study, we investigate the efficacy of estimating finger force from integrated soft sensors and use it to estimate contact states. We use a neural network for force regression, collecting labelled data with a force/torque sensor and a range of test objects. Subsequently, we apply this model in a plug-in task scenario and demonstrate its validity in estimating contact states.
\end{abstract}

\section{Introduction}
Soft fingers have the advantage of intrinsic adaptability to the grasped object's geometry \cite{shintake2018}, supporting power or precision grasps in tasks where objects vary or are sensitive. While soft fingers can enable grasping of new objects \cite{billard2019} and passive compensation of alignment error \cite{hartisch2023}, new challenges emerge, such as slip \cite{taylor2022}, limited force capacity \cite{liu2021, park2020}, loss of accuracy \cite{li2021a}, and grasp stability \cite{stabile2022,jang2023soft}. 

\begin{figure}[t]
\centering
    \subfloat[Tactile signal processing]{\includegraphics[width=\columnwidth]{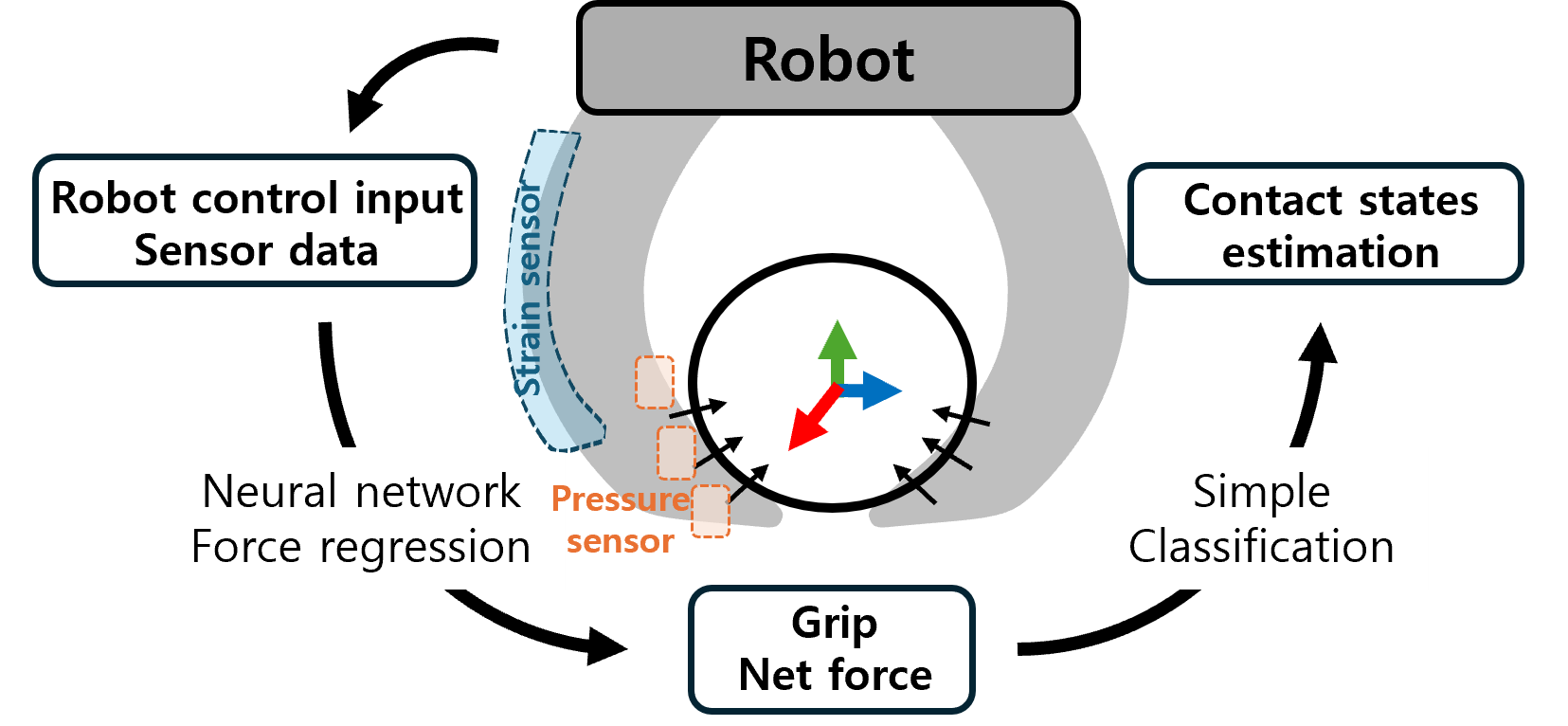}}  \\
    \subfloat[Example contact states]{\includegraphics[width=0.75\columnwidth]{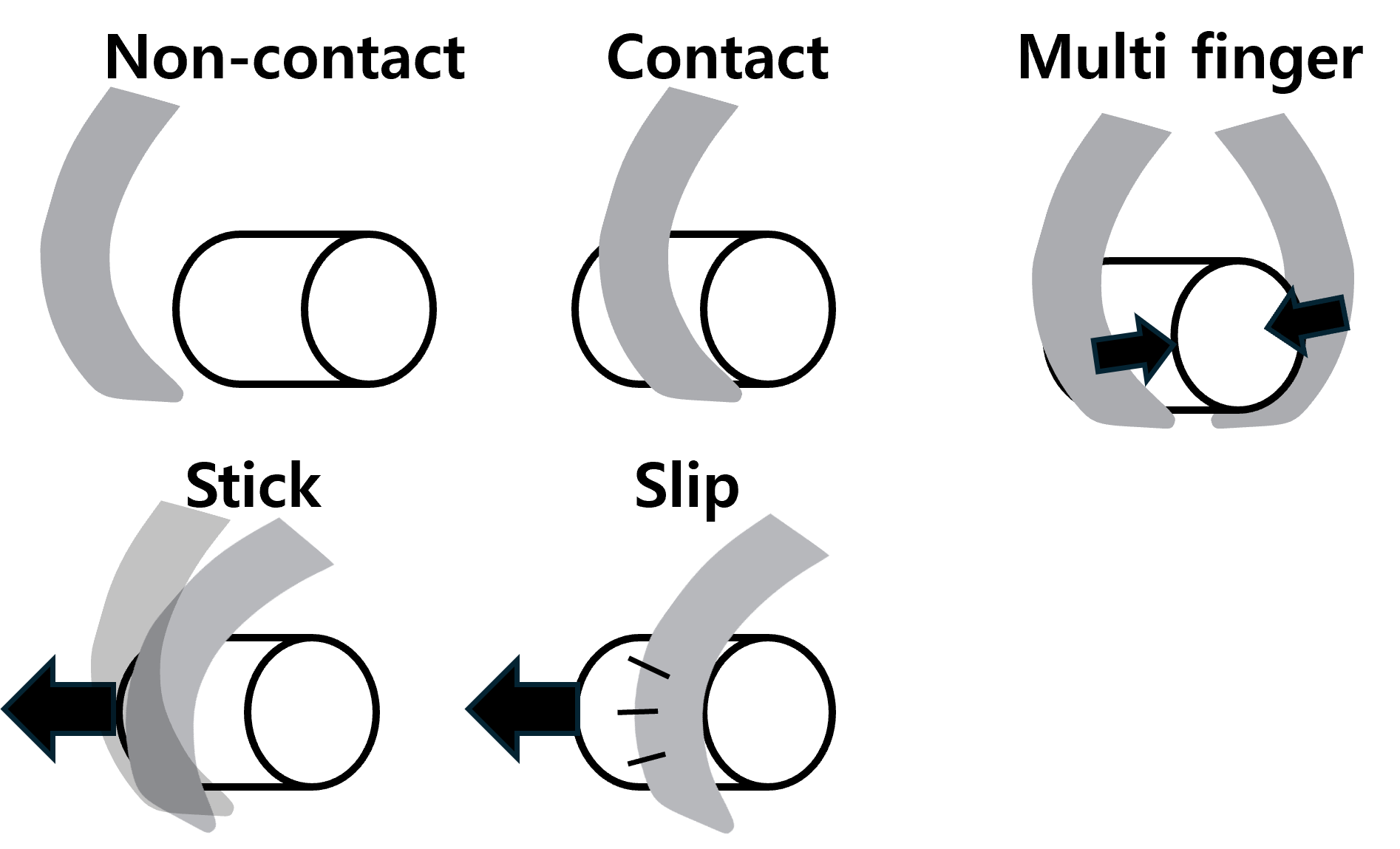}}
\caption{Soft fingers (a) with integrated soft pressure and strain sensors grasping an object, and (b) example internal grasp states which can be detected based on the total finger force.}
\label{fig:schematic}
\end{figure}

Mechanical design can improve some aspects of soft fingers, e.g. increase force capacity with new actuation methods \cite{azami2023} or increased actuated degrees of freedom \cite{teeple2020}. However, sensors provide an alternative approach, where tactile sensors can provide task state information \cite{tian2019a, xu2024}, estimate grasp quality \cite{li2014a, park2020}, or directly estimate the pose of a grasped object \cite{li2021a}. 

Tactile sensing with machine learning enables real-time identification of contact states such as external forces \cite{kim2023, debarrie2021, suh2021}, slip \cite{taylor2022}, and contact location \cite{massari2020}. This contact information can be internal forces in the grip (i.e. individual forces between finger and grasped object), e.g. to evaluate grasp quality \cite{dong2019}. The contact information can also be extrinsic, i.e. the total forces between  grasped object and environment. Extrinsic contact force information is imperative for detecting and correcting deviations in contact-rich manipulation, for example estimation of the contact mode in assembly tasks \cite{pankert2023, haninger2022b} or triggering of contact primitives \cite{hogan2020}.  

High-resolution tactile sensors, such as image-based, can be used to estimate the pose \cite{taylor2022, villalonga2021}. They can also be used to estimate properties of the grasped object, such as hardness or texture.  Integrated strain sensors can be used to estimate the position and 1 degree of freedom external forces at the fingertip \cite{thuruthel2019}. Detection of object slip can be done with either image-based \cite{taylor2022} or strain-sensor-based \cite{huang2022a} sensors. Simultaneous estimation of force and point of application has been realized with fiber Bragg grating sensors \cite{massari2020}, but not yet shown integrated to soft fingers. Force estimation can also be realized by visual deformation of a soft structure \cite{zhang2018a} or soft grasping surfaces \cite{suh2021}, provided an estimated stiffness of the soft structure is known.  

Tactile sensors can also provide insight into the internal forces in a grasping process.  These internal forces are critical for evaluating grip stability and slip \cite{roa2015, dong2019}, which are especially important in precision grasps \cite{teeple2020}. The grasp internal forces can be used to find motions which stick or slip \cite{chavan-dafle2019} and the force capacity before slip occurs \cite{park2020}. 

We propose a machine-learning method for estimating the total force applied by a soft finger from integrated pressure and strain sensors as seen in Figure \ref{fig:schematic}. The estimator is trained from labelled data collected on a range of object. Compared with \cite{thuruthel2019}, our approach does not use an external camera system and considers grasp forces along the finger, not just the finger tip. Compared with \cite{park2020}, this approach includes strain sensors and uses regression with a neural network to estimate the finger force. The approach is analyzed in generalizability to novel objects, ablation studies for sensor modalities, and applied to a contact-rich assembly task. 

\section{Soft finger contact state characterization}
This section introduces models for the forces and pressures in a soft finger grasping problem.

\subsection{Soft finger grasping model}
\begin{figure}[h]
    \centering
    \includegraphics[width=0.8\columnwidth]{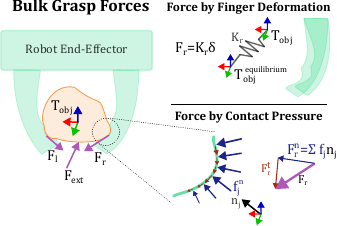}
    \caption{Grasping force model, where each finger exerts a total force on the grasped object. The total forces can be measured by either the bulk deformation of the finger from an equilibrium position (right, top), or the total integral of surface pressure (right, bottom).}
    \label{fig:force_model}
\end{figure}

We consider that the $i^{\mathrm{th}}$ finger exerts a total force vector $F_i$ on a grasped object, as shown in the planar case in Figure \ref{fig:force_model}. The individual finger forces sum to the total external force $F_{ext}=\sum_i F_i$. This external force is the gravitational and inertial forces of the object, as well as any external contact force applied to the object. 
%\begin{figure}
%    \includegraphics[width=0.8\columnwidth]{New_fig/Chapter2/3.diagram.png}
%\caption{Internal and external forces during grasping}
%\label{fig:contact_states}
%\end{figure}

An individual finger's force is composed of a normal pressure profile over a local contact patch and a tangential friction force, $F_i = F^n_i + F^t_i$. The normal force can be estimated from pressure measurements at the surface as $F^n_j = \sum_{j\in \mathcal{J}_i} f^n_j n_j$, where $\mathcal{J}_i$ is the set of pressure sensors on the finger, $f^n_j\in\mathbb{R}$ the normal pressure, and $n_j \in \mathbb{R}^3$ is the contact normal. 

As the fingers are soft, the total finger force also results in a deformation from the equilibrium position. If we consider a linearized stiffness model, this force is $F_i=K_i\delta_i$, where $K_i$ is a linearized stiffness and $\delta_i$ the deformation from equilibrium. If strain sensors in the finger allow measurement of the deformation $\delta_i$, this could also inform an estimate of the finger force.

These two complementary force measurement methods -- bulk deformation and surface pressure -- involve corresponding unknowns of finger stiffness $K_i$ and contact normal orientation $n_i$. These characteristics will depend on the contact conditions with the object and the finger form, which can be correspondingly estimated by pressure and strain sensors.  This motivates the use of a regression model which takes both strain and surface pressure to detect total force.

\subsection{Internal grasp states} 

To check the utility of the simplified bulk force model in Figure \ref{fig:force_model}, we experimentally investigate the accuracy of a simplified Coulomb model for slip at the bulk force level. We investigate Coulomb friction because slip is an important grasp state we would like to be able to predict or detect from the force estimation system.  We test a range of finger conditions, materials, test objects, and robot positions. 
\begin{figure}[h]
    \centering
    \includegraphics[width=0.6\columnwidth]{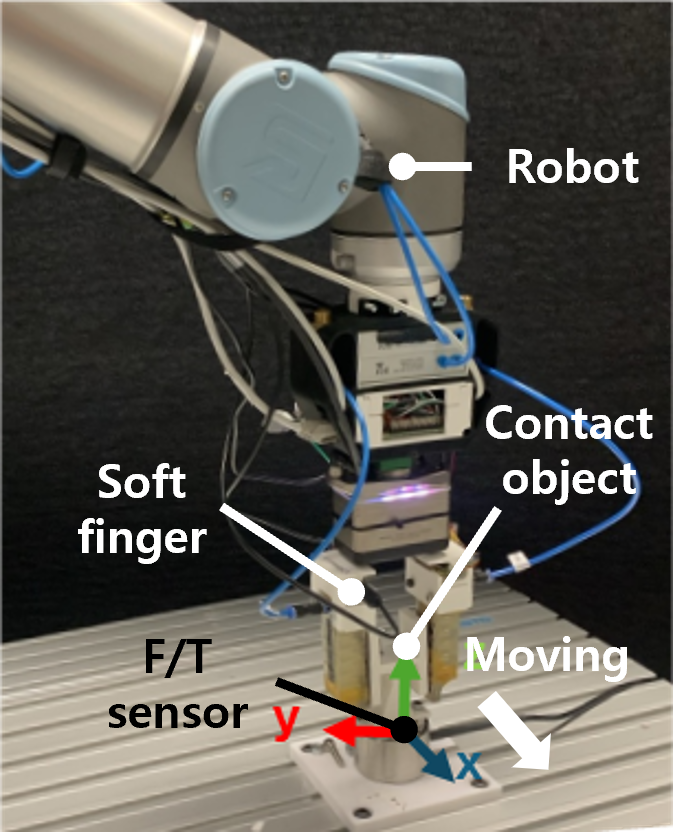}
    \caption{Experimental setup for internal grasp state experiments}
    \label{fig:exp_setup}
\end{figure}
\begin{figure*}[t!]
   \centering
    \includegraphics[width=1.8\columnwidth]{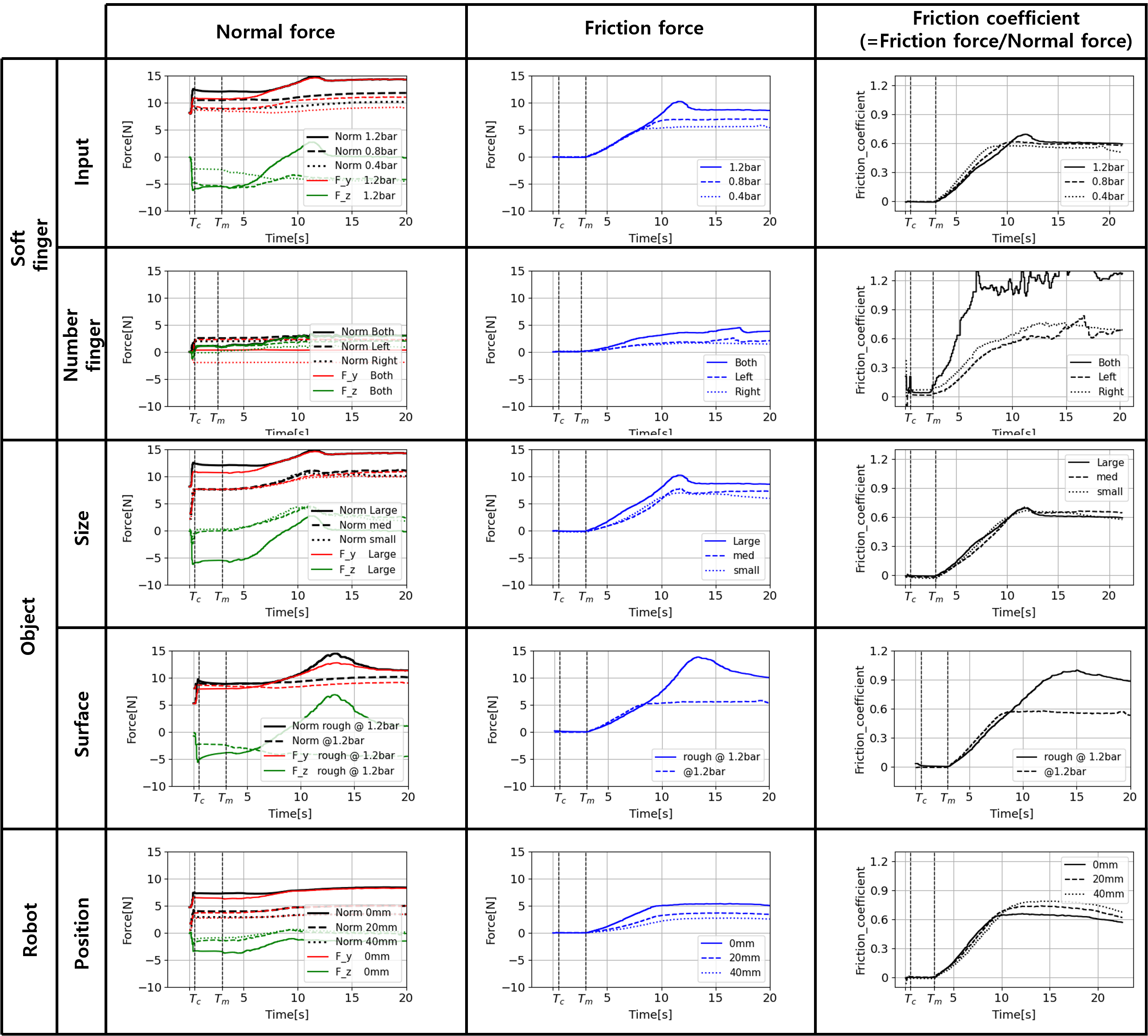}
\caption{We validate the ability to use internal forces for soft finger slip. The object size, robot offset and material affect the normal and cross-forces, and the stick, micro-slip and sliding can be distinguished.  The results indicate (1) the Coulomb model provides 20\% error, showing measuring normal force is useful for predicting slip, (2) the friction model also applies to two-finger applications.}
\label{fig:force_exp}
\end{figure*}
%% Hun Experiment analysis
The experimental setup, depicted in Figure \ref{fig:exp_setup}, investigates the key factors influencing internal grasping, including soft finger parameters (input pressure, number of fingers), robot positioning (relative finger-object position), and object characteristics (size, surface properties). Soft fingers, made from Silicone RT625 and based on PneuNets\cite{polygerinos2015a}, are mounted on a Universal Robots UR10, with input pressures controlled by a valve controller (VEAB-L-26-D9-Q4-V1-1R1, Festo). Experiments were conducted using both single and dual fingers, with the dual-finger setup aligning the center of the fingers with the object's center, maintaining a 30 mm distance between them. The robot adjusts the horizontal position of the fingers relative to the object. The object, 3D printed from PLA with a 30 mm radius, has a modified arc length to ensure full contact, and for one experiment frictional tape is applied to change the friction characteristics.

To measure internal grip force under each parameter set, the test object is fixed to a Force/Torque sensor (ME-Meßsystem MP11, 500N, 20Nm, 125Hz). The procedure involves actuating the soft finger, making contact ($T_c$), and moving the robot 2 seconds later ($T_m$) at $2$ mm/s in $+x$ over $30$ mm. As the robot moves, the friction force $F_f=F_x$ is measured, corresponding to the force aligned with the direction of movement. The normal force is calculated using the measured data orthogonal to the movement direction, specifically $F_n = \sqrt{F_y^2+F_z^2}$. The friction coefficient is then derived from these measurements as $F_f/F_n$. Each experiment is conducted three times for each set of parameters, with the mean values plotted in Figure \ref{fig:force_exp} to ensure accuracy and reliability of the results.

%1. Contact Initiation and Transition States: (How to estimate contact state )
Contact initiation is easily distinguishable by observing the onset of contact, marked by $T_c$ in the figure of normal force. The transition between slip and stick states involves distinct dynamics, with friction forces exhibiting noticeable fluctuation post-contact, marked by $T_m$ in the figure of normal force. Minimal variations in normal forces are observed compared to friction forces during the stick-slip states, and friction forces progressively increase until complete slippage occurs. Distinguishable patterns in friction coefficients indicate different contact states.

%2. Parameter influence on Contact dynamics
Normal force is largely affected by robot position and object size, as well as finger pressure. When the robot is moving but before slip occurs, the rate of increase in friction force can be attributed to the finger's stiffness $K_i$. We observe that this stiffness is affected by changes in position and object size, but not finger pressure and surface characteristics. In the slip condition in the final phase of the trajectory, the effective Coulomb friction coefficient is largely consistent just over $0.6$, except when friction tape is applied to the object and when two fingers are used. 
%This suggests that with known materials, predicting the total normal force allows a good estimate of the maximum friction force.
This suggests that, with known materials, grasping force in both normal and transverse direction can provide a reliable estimation of contact states. However, in multi-finger grasping, the normal force is compromised, resulting in a reduction of its magnitude, while the frictional force is multiplied. This effect makes it challenging to estimate contact states using these methods.

\begin{align*}
    \begin{split}
    \text{ Contact state } 
                   & =\begin{cases}
                        %Noncontacct\hspace{0.5cm} : \sqrt{F_n^2+F_f^2}=0\\ 
                        Noncontacct\hspace{0.5cm} : F_n=0\\ 
                        %Contact\hspace{1.2cm}              : \sqrt{F_n^2+F_f^2}>0\\
                        Contacct\hspace{1.1cm} : F_n>0\\ 
                        Stick \hspace{1.6cm}:  F_f/F_n <0.6\\
                        Slip \hspace{1.8cm}:  F_f/F_n >0.6\\                    
                      \end{cases}.
    \end{split}
\end{align*}

\section{Force Regression}
%The previous section showed that finger normal force can provide good estimates for slip force, and this section shows how we can estimate this force.
The previous section showed that finger grip force, which is normal and transverse force, can provide good estimates for contact states and this section shows how we can estimate these forces.
To address the challenges faced by soft sensors, such as hysteresis, non-linearity, and crosstalk, a strategy involving data-based modeling is often used. This approach focuses on collecting a wide range of data to help build a model that can handle these issues.

\subsection{Soft gripper systems with soft sensors}
\begin{figure}[h]
    \centering
    \subfloat[Pressure and strain sensor]{\includegraphics[width=0.5\columnwidth]{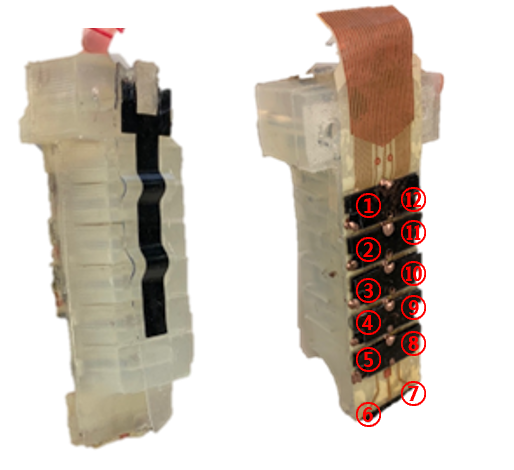}}
    \label{fig:soft_sensor}
    
    \subfloat[Strain sensor verification]{\includegraphics[width=0.47\columnwidth]{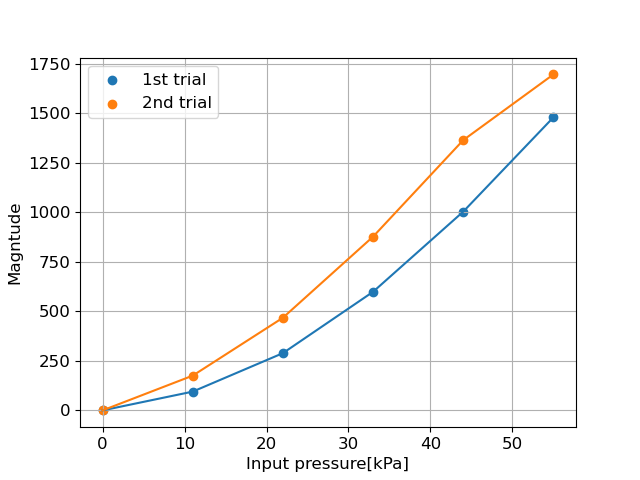}}
    \subfloat[Pressure sensor verification(3rd, 4th, 8th, 9th points are pressed)]{\includegraphics[width=0.47\columnwidth]{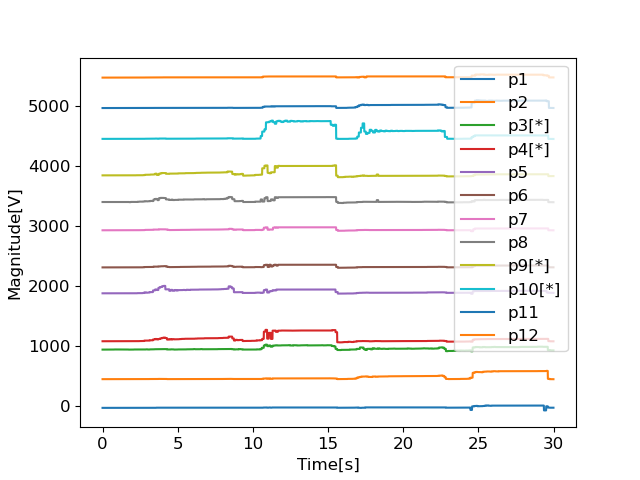}}
\caption{Soft gripper system with soft sensors (top) as well as characterization of sensors in simplified tasks (bottom). }
\label{fig:sensor_integration}
\end{figure}
The soft gripper system from Figure \ref{fig:exp_setup} includes two pneumatic fingers, each with flexible sensors—-one strain sensor and twelve pressure sensors per finger—-as seen in Figure \ref{fig:sensor_integration}(a). The sensors are dielectric elastomer sensors, fabricated for sensing compression \cite{bose2014} and strain \cite{obrien2014}, respectively. The capacitance is measured with a Teensy LC, and a Raspberry Pi 4 publishes the sensor values at a rate of 100 Hz.

In Figure \ref{fig:sensor_integration}(b), the performance of the strain sensor is illustrated. A notable observation here is the presence of hysteresis – a phenomenon where the sensor does not return to its original state after being deformed. Additionally, the sensor displays some non-linearity in its response, making its readings less straightforward to interpret.

Further complications are observed in the pressure sensor verification, as shown in Figure \ref{fig:sensor_integration}(c). When specific sensing points (namely the 3rd, 4th, 9th, and 10th points from Figure \ref{fig:sensor_integration}(a)) are pressed, the sensors at those locations can detect the pressure. However, this pressure also inadvertently affects the readings from adjacent sensors, a phenomenon known as cross talk. This effect means that even unpressed sensors show changes in their signals, complicating the interpretation of the data.

These issues – hysteresis, non-linearity, and cross talk – pose challenges in using the signals from the soft sensors directly. It indicates the need for sophisticated data processing or calibration techniques to accurately interpret the sensor data.

\subsection{Data collection method}

\begin{figure}[h]
    \centering
    \includegraphics[width=0.8\columnwidth]{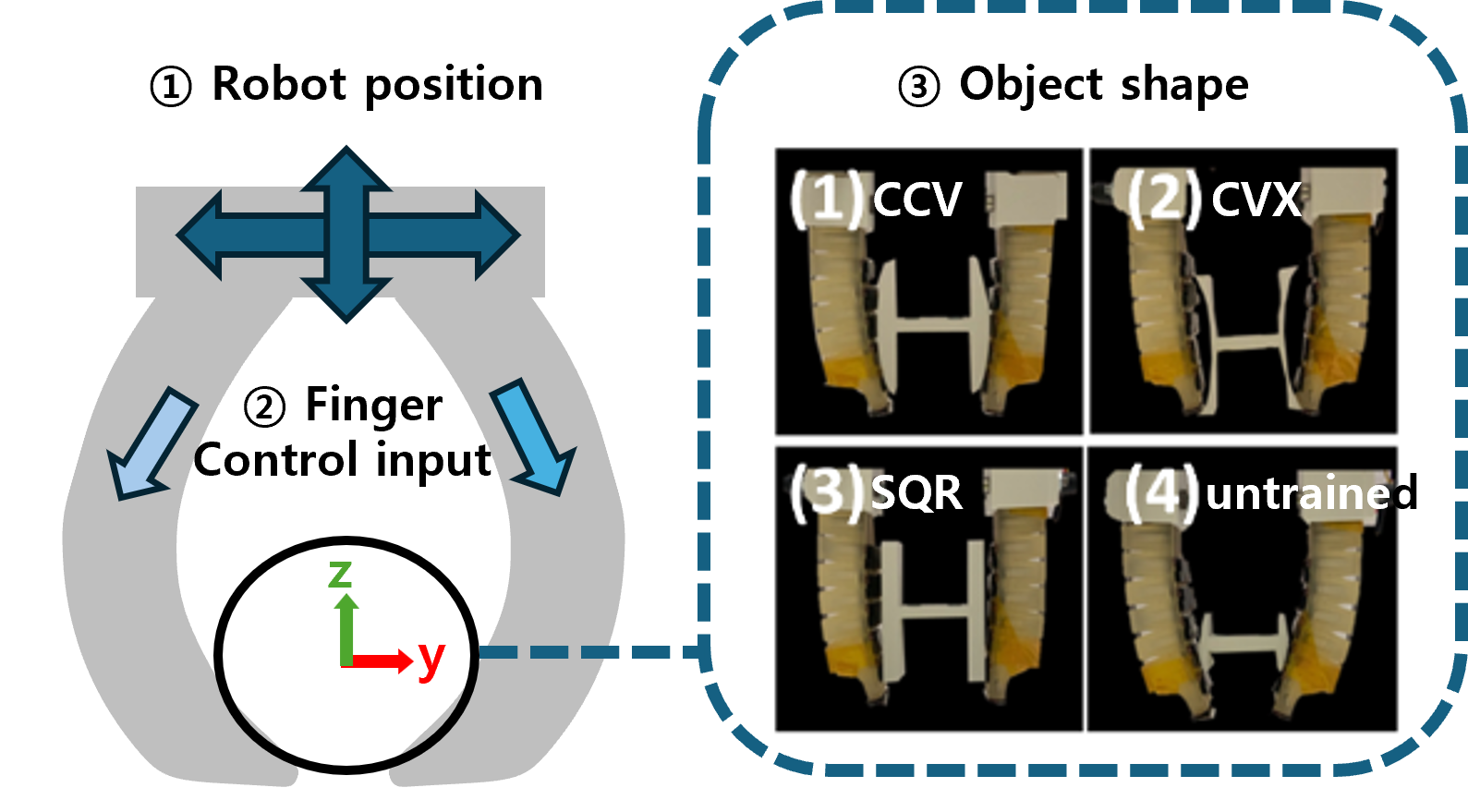}
    \caption{Parameters varied for the data collection. The object $(4)$ is used for validation, not included to the training data.}
    \label{fig:datasetup}
\end{figure}

The setup for gathering this data is the same as shown in Figure \ref{fig:exp_setup}. It's important to have a variety of conditions in the data, so different-shaped objects—convex, concave, and square—are used in the tests, as seen in the right of Figure \ref{fig:datasetup}. Among these, different sizes of convex objects are included, with the smallest being used only for validation. This means it's not included in the training phase but is used later to check how well the model works with an object it hasn't encountered before.

To test how well the model can generalize, different pressures, from 0 to 60 kPa, are applied to each finger. These grasps are made at a wide range of robot positions, with the robot moving in Y and Z directions. Each test is repeated four times, with the soft gripper following the same path and pressure. Out of these, three repetitions are used to train the model, and the last one is for verifying how well the model can predict under new conditions.

\subsection{Network and training}
\begin{figure}[h]
    \centering
    \includegraphics[width=1\columnwidth]{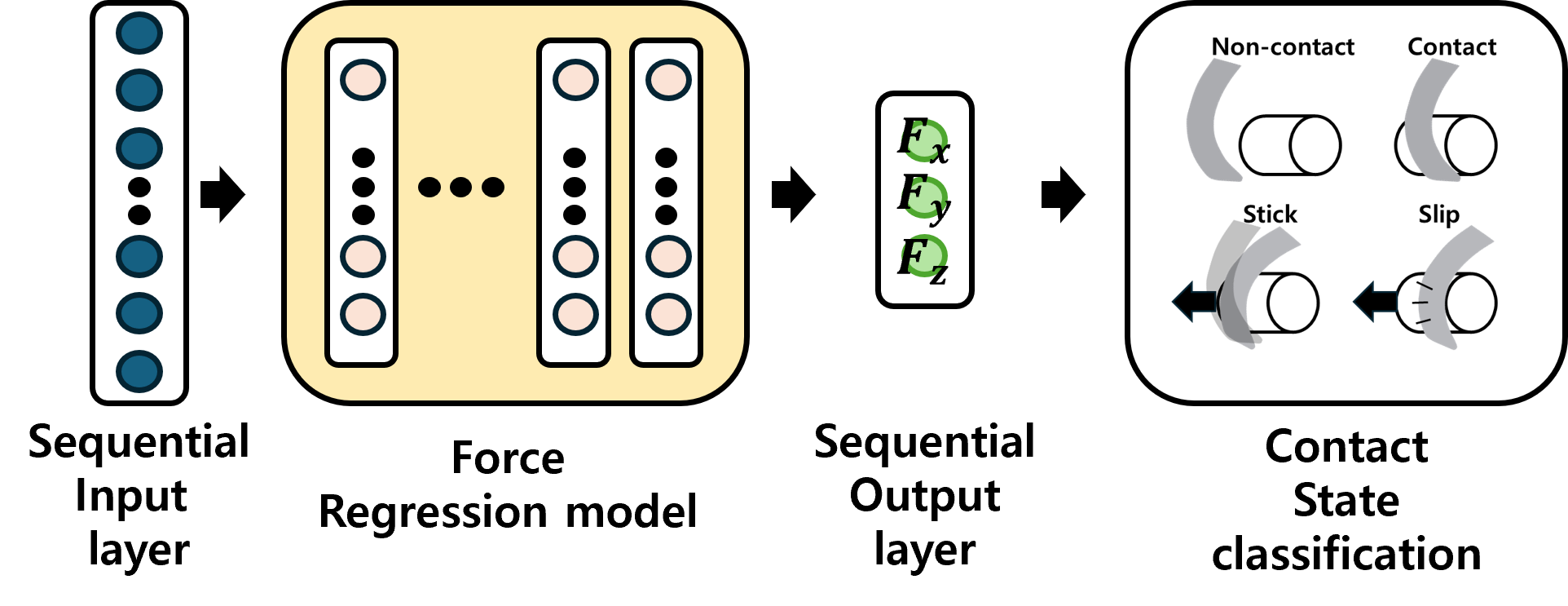}
    \caption{Data pipeline from input through the contact state detection}
    \label{fig:layer}
\end{figure}
Deep learning model hyperparameter tuning is helpful to better understand how soft sensors affect force regression. We compared three different types of deep learning models, each with varying layers and hidden nodes. To keep things consistent, we used a batch size of 32, ran 50 epochs, and employed the ADAM optimizer for our models.

Before we started any deep learning processes, all the sensor data is scaled to normalize. This step helps to compensate long-term drift in the sensor offset. We used Scikit-Learn’s robust scaler in our data preprocessing because it handles outliers in the soft sensor data well \cite{sklearn_api}. 

We also compared against recurrent models, which can consider time-dependent variables such as drift or hysteresis, but they can also make the model more complex. This complexity might lead to issues like over fitting or the need for extensive data collection.

We used four different model architectures, a multi-layer perceptron, recurrent network, long short-term memory and gated recurrent network (MLP, RNN, LSTM, GRU) and experimented with three varying numbers of layers (1, 5, 10) and hidden nodes (1, 5, 10). 

The choice of input features was also a critical factor in our model’s performance. We tested seven different combinations of input features to see which one yielded the best results. These combinations included:  input pressure alone(t1), only strain sensors(t2), only pressure sensors(t3), input pressure with strain sensors(t4),  input pressure with pressure sensors(t5), strain sensors with pressure sensors(t6), and input pressure with all sensors(t7). We maintained the same model parameters for each of these combinations to ensure a fair comparison.

\begin{comment}
Commented code

\begin{table}[h!]
    \centering
    \begin{tabular}{|r|l|}
        \hline
        \textbf{Parameter name} & \textbf{Value} \\
        \hline\hline
        \textbf{Model} & MLP, RNN, LSTM, GRU \\
        \hline
        \textbf{Layer } & 1, 5, 19 \\
        \hline
        \textbf{Hidden Node } & 1, 5, 10 \\
        \hline
        \textbf{Input feature[\(type\)] } & 1,2,3,4,5,6,7  \\
        \hline
        \textbf{Object} & cvx, ccv, sqr, untrained \\
        \hline
    \end{tabular}
        \caption{Parameters for deep learning\label{tab:Parameter_experiment}}

\end{table}
\end{comment}

\begin{table}[]
\centering
    \begin{tabular}{|cc|c|}
        \hline
        \multicolumn{2}{|c|}{\textbf{Parameter}}      & \textbf{Value}         \\ \hline
        \multicolumn{1}{|c|}{\multirow{3}{*}{\textbf{\begin{tabular}[c]{@{}c@{}}Model\\ Type\end{tabular}}}} & \textbf{Model} & MLP,RNN,LSTM,GRU           \\ \cline{2-3} 
        \multicolumn{1}{|c|}{} & \textbf{Layer}       & 1, 5, 20               \\ \cline{2-3} 
        \multicolumn{1}{|c|}{} & \textbf{Hidden Node} & 1, 5, 10               \\ \hline
        \multicolumn{2}{|c|}{\textbf{Input feature{[}type{]}}}                                                                & t1, t2, t3, t4, t5, t6, t7 \\ \hline
        \multicolumn{2}{|c|}{\textbf{Object}}         & CCV,CVX,SQR, untrained \\ \hline
    \end{tabular}
    \caption{Hyper parameter study}
    \label{Table:ta}
\end{table}

% Network architecture 
% Training process

\section{Force Estimation Validation}

To validate our model, we compared the Root Mean Square Error (RMSE) of force prediction for each type of object.

\subsection{Model parameter comparison}
\begin{figure}
    \centering
    \includegraphics[trim={0cm 0.3cm 0cm 0cm}, clip,width=0.8\columnwidth]{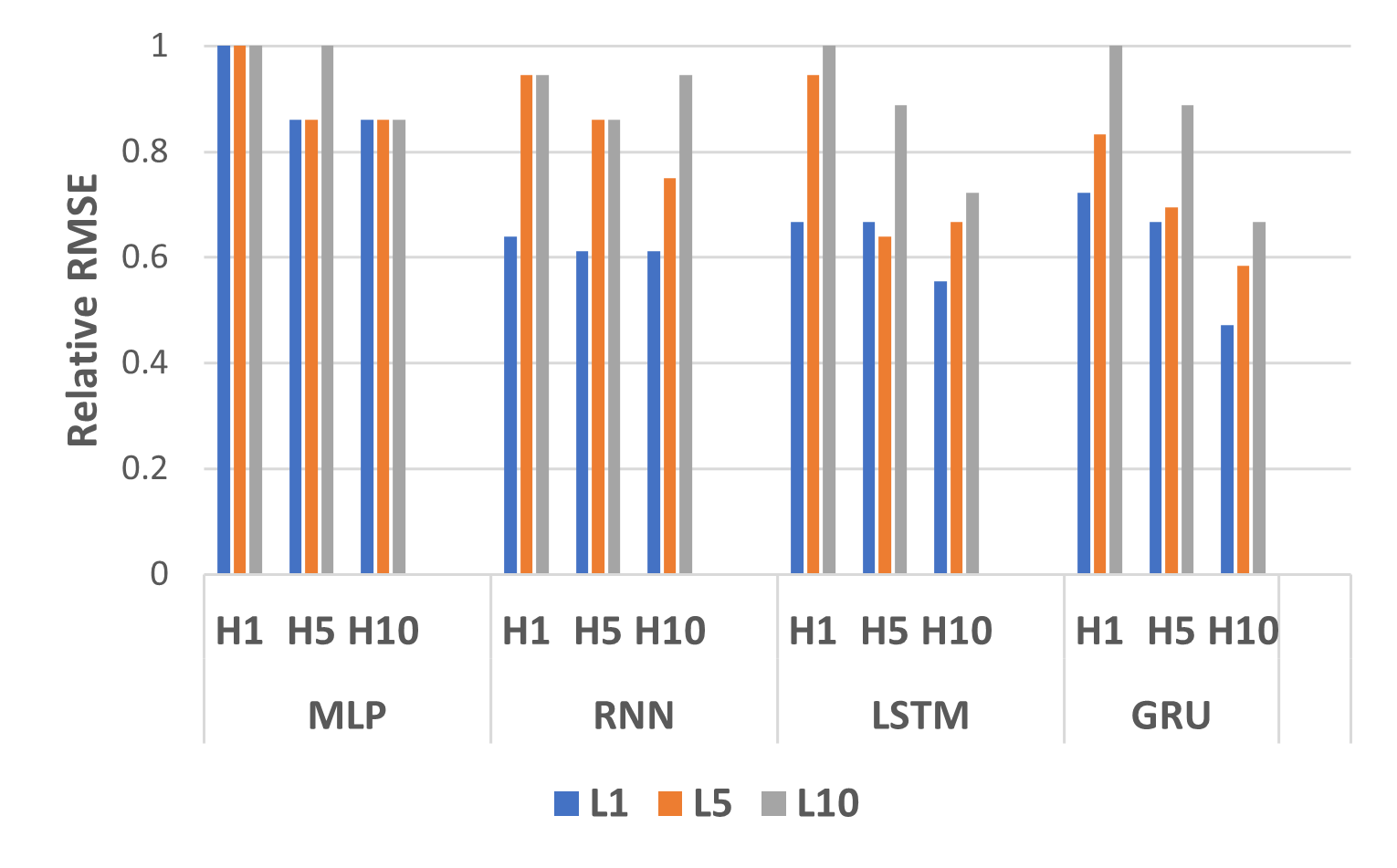}
    \caption{Relative RMSE results of different parameter}
    \label{fig:exp_1}
\end{figure}

In Figure \ref{fig:exp_1}, we see the outcomes of various models and parameters, all adjusted relative to each other by normalizing them against the highest RMSE result ($0.36$). The data indicates that simply increasing the number of layers in a model doesn't necessarily enhance its performance. In fact, too many layers can complicate the model without improving its estimations. However, having a higher number of hidden nodes generally leads to better results. Notably, the MLP model doesn't show significant improvement compared to other models. This is likely due to the absence of a recurrent term in MLPs, which limits their predictive capabilities.

Among all the models tested, the GRU model with one layer and 10 hidden nodes emerged as the most effective. GRU models are simpler than LSTM models as they don't need as many hidden nodes or layers. Their structure also helps them overcome the vanishing gradient problem, a common issue in traditional RNNs. Therefore, the GRU model with one layer and 10 hidden nodes is used for following work.

%\begin{figure*}[h]
%    \centering
%    \includegraphics[width=1\columnwidth]{figs/4. Verification/exp_2_best.png}
%    \caption{Best accuracy(GRU, layer=1, Hidden node=10}
%    \label{fig:best}
%\end{figure*}

\begin{figure*}[!t]
    \centering
    \includegraphics[trim={0cm 2cm 0cm 2cm}, clip, width=0.8\textwidth]{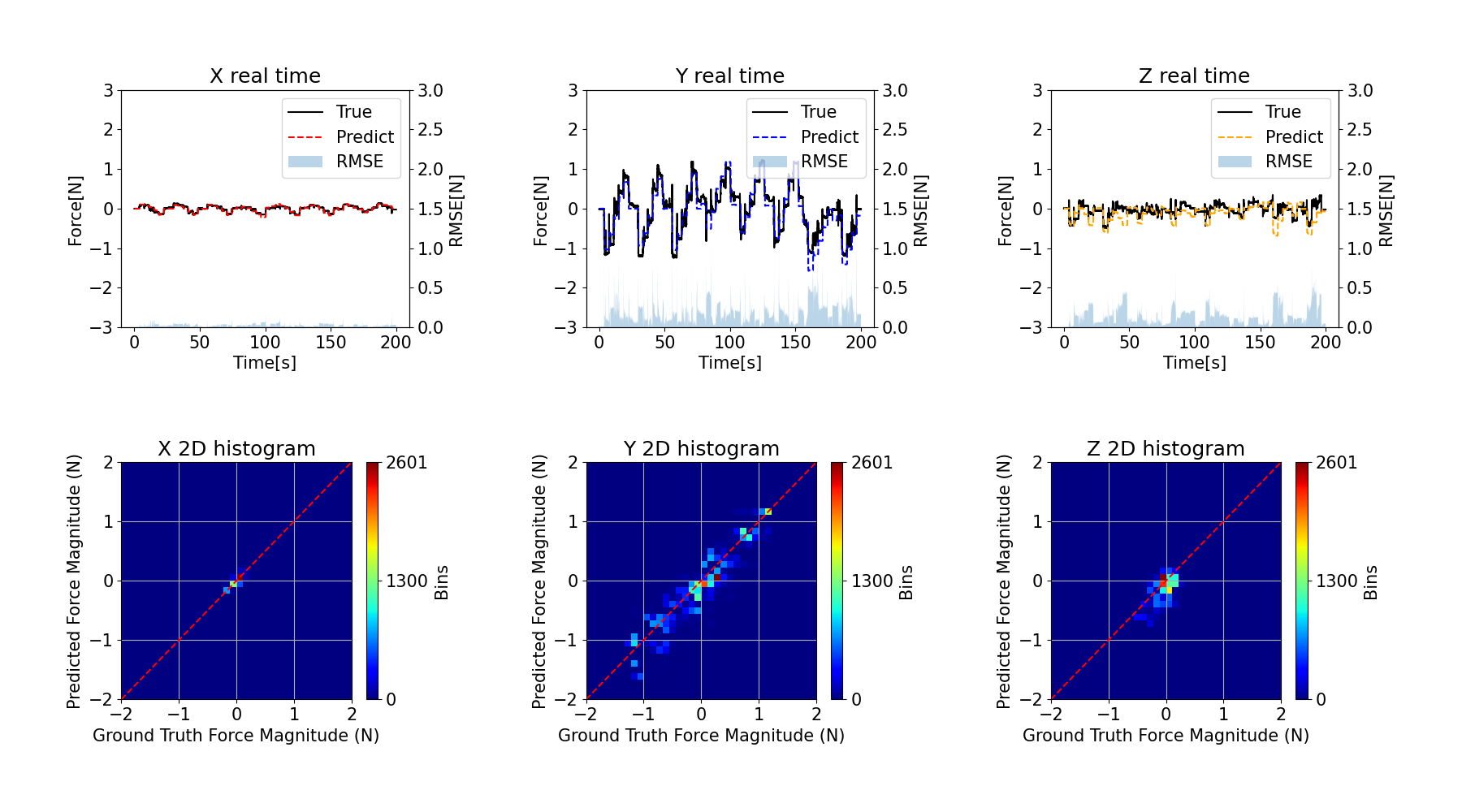}
    \caption{Force regression results in time and histogram for a GRU with 1 layer and 10 hidden units, displayed over the force dimensions.}
    \label{fig:best}
\end{figure*}

Figure \ref{fig:best} shows the regression results with the GRU model (1 layer, 10 hidden nodes). It can be seen that $F_y$ has a larger range of force, this is the main grasp force direction and also the highest sensitivity of both pressure and strain sensors.  Estimation of $F_z$ has comparable RMSE over a much smaller force range.

\subsection{Ablation of sensing modalities}
\begin{figure}[H]
    \centering
    \includegraphics[trim={0cm 1cm 0cm 0cm}, clip,width=0.8\columnwidth]{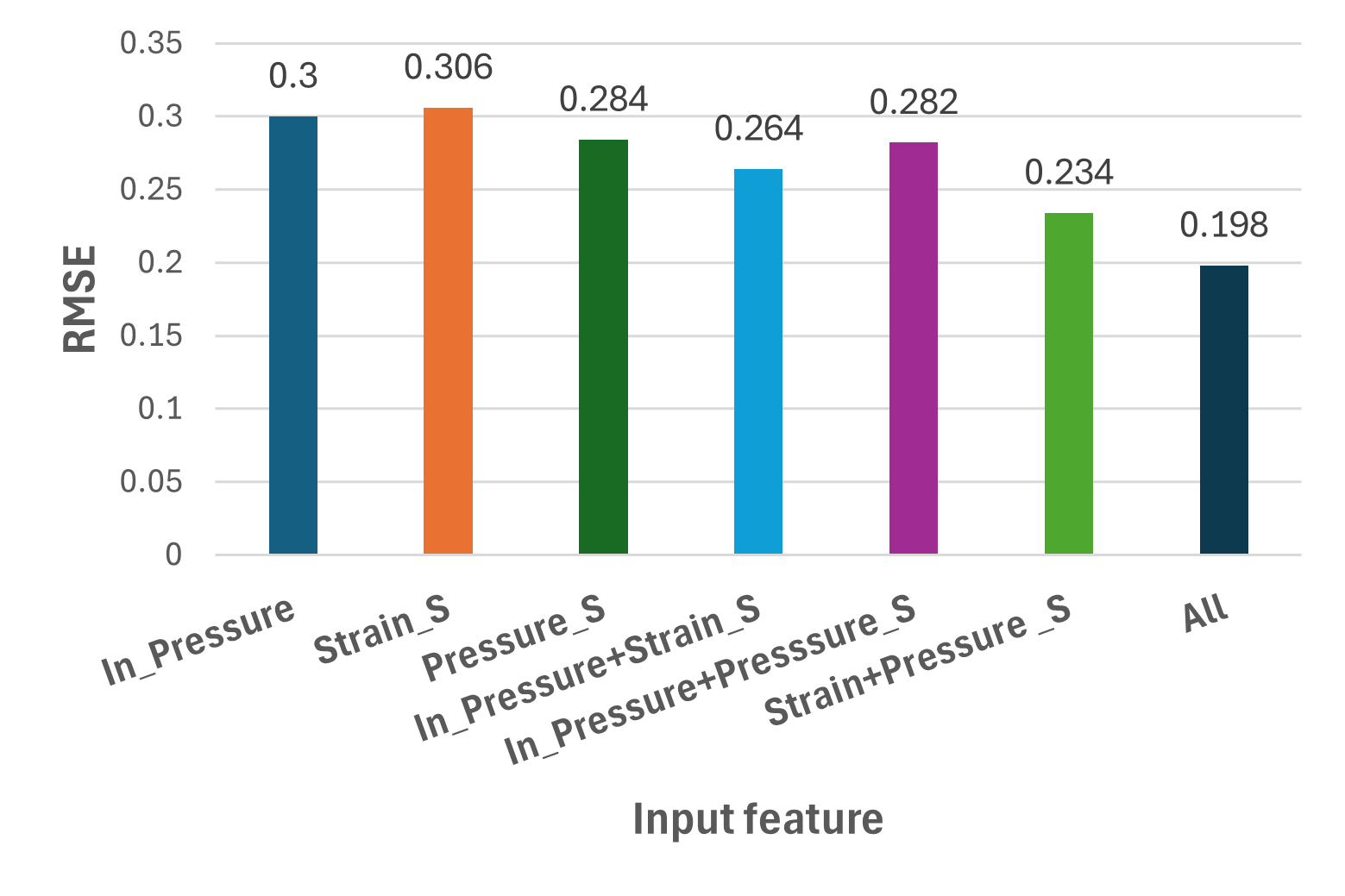}
    \caption{Comparison of input features effect on validation RMSE}
    \label{fig:feature_comparison}
\end{figure}

When comparing different input features with the same model and parameters, it can be seen in Figure \ref{fig:feature_comparison} that more input features generally lead to more accurate results. Even though the combination of input pressure and pressure sensors and the combination of strain and pressure sensors have the same number of input features, the latter performs better. This suggests that despite their non-linearity and hysteresis characteristics, strain sensors provide valuable information.

\subsection{Generalizability over objects}
\begin{figure}[H]
    \centering
    \includegraphics[trim={0cm 1cm 0cm 0.5cm}, clip, width=0.6\columnwidth]{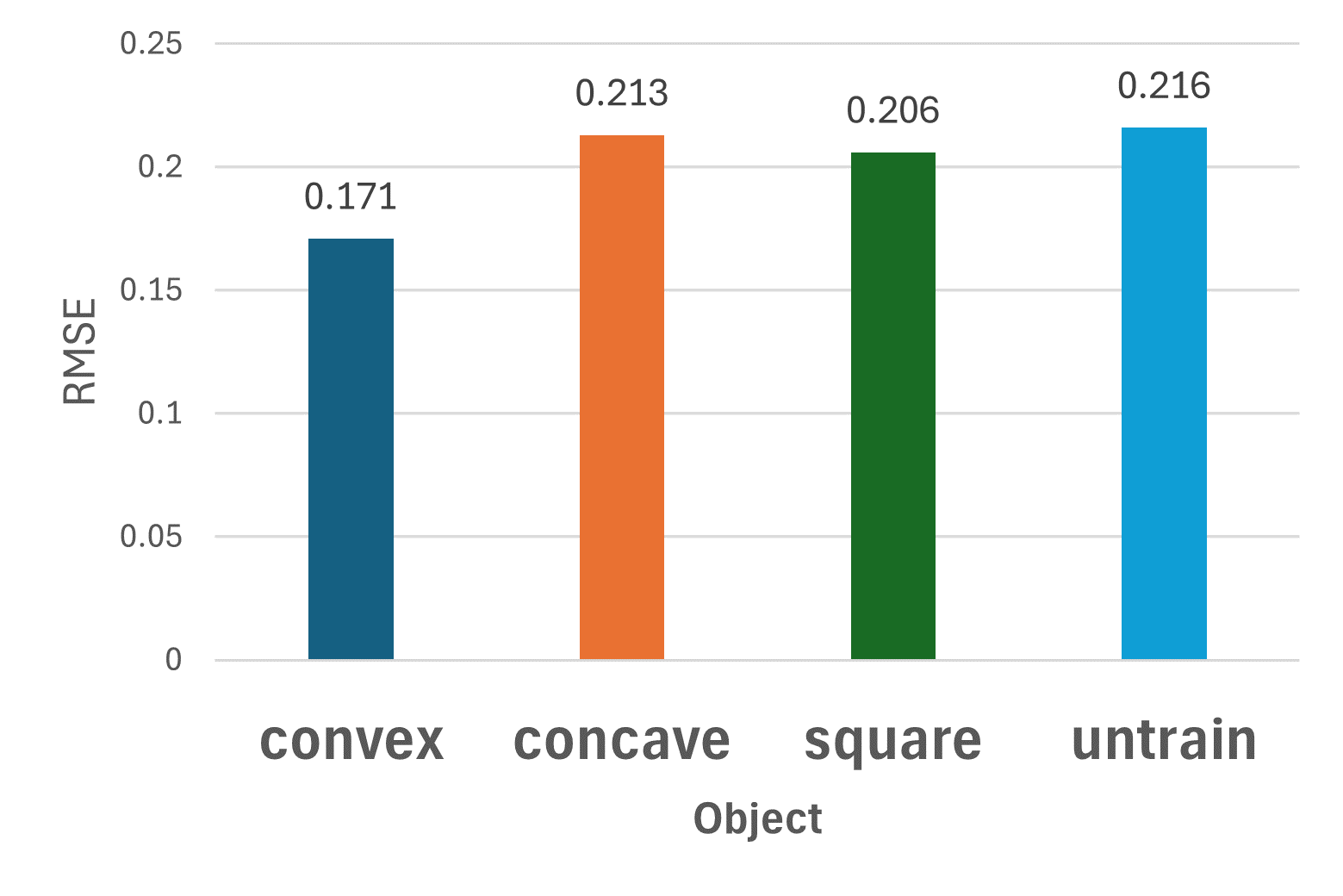}
    \caption{RMSE over different object types, with the untrained object not being in the training data set, (4) from Figure \ref{fig:datasetup} }
    \label{fig:object_comparison}
\end{figure}
Figure \ref{fig:object_comparison} shows the object force estimation results using the same input features and model. The model performs well with conformally deforming soft fingers around cylinder-shaped objects. Interestingly, even for an untrained shape, a smaller circular object, the model estimates forces comparatively well, showing little difference from its performance with convex shapes. This indicates the model's robustness and adaptability in various scenarios, making it a promising tool for applications involving complex and varied tactile interactions.

\section{Applications to robotic manipulation}
In the context of manipulation, slip detection and contact mode identification are key functionalities which can be enabled by force estimates.

\subsection{Slip detection}
\begin{figure}[h]
    \centering
    \includegraphics[ width=0.8\columnwidth]{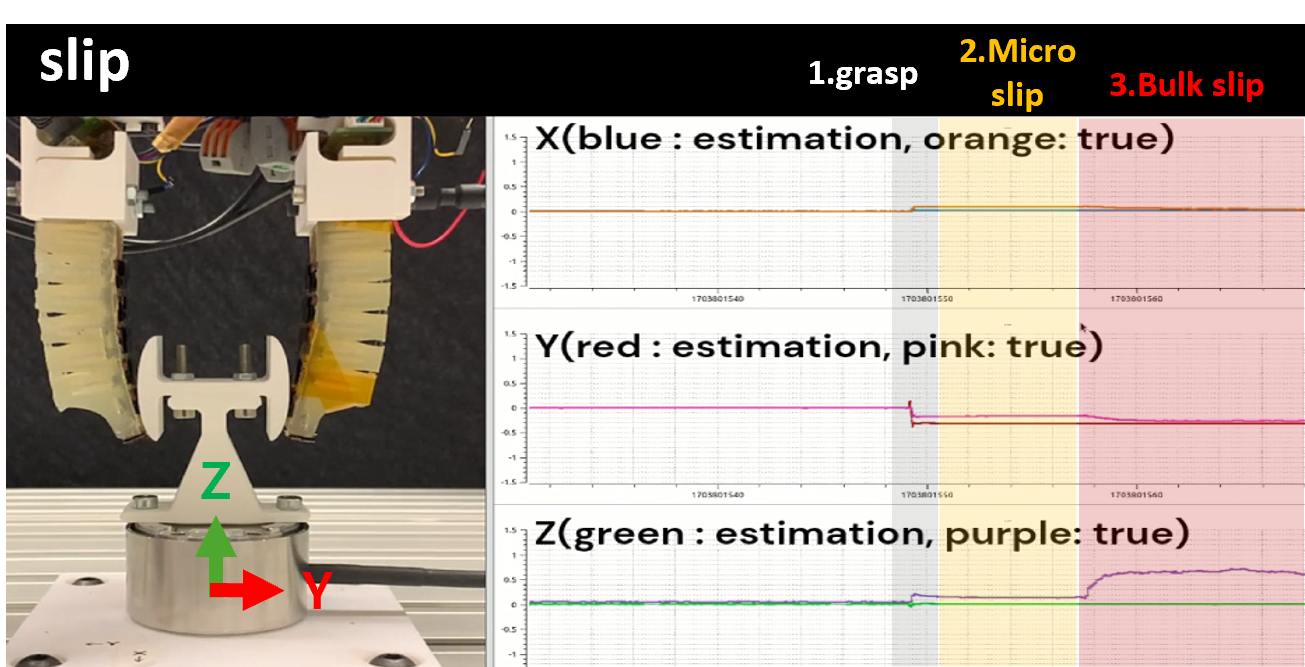}
    \caption{Finger slip experiment setup and estimated forces}
    \label{fig:validate_slip}
\end{figure}
The first application scenario entails slip detection, tested in manipulation of a curved object. real-time force estimation, as illustrated in Figure \ref{fig:validate_slip}, shows predictive force profiles alongside actual force data derived from a Force/Torque (F/T) sensor. 

With the object grasped, the robot moves downwards, prompting slip discernible in two phases: micro slip and bulk slip. During micro slip, minimal alterations in both prediction and F/T sensor data are observed, owing to sustained contact adherence. Conversely, bulk slip manifests with pronounced shifts in F/T sensor data, indicative of slip onset. The predictive model detects the Y forces during grasp, but does not detect the change during bulk slip.  We attribute this to the limited sensor sensitivity in this direction: shear forces are not measured by the pressure sensors, and the fingers are not highly deformable in this direction.

\subsection{Contact detection}
\begin{figure}
    \centering
    \includegraphics[width=0.8\columnwidth]{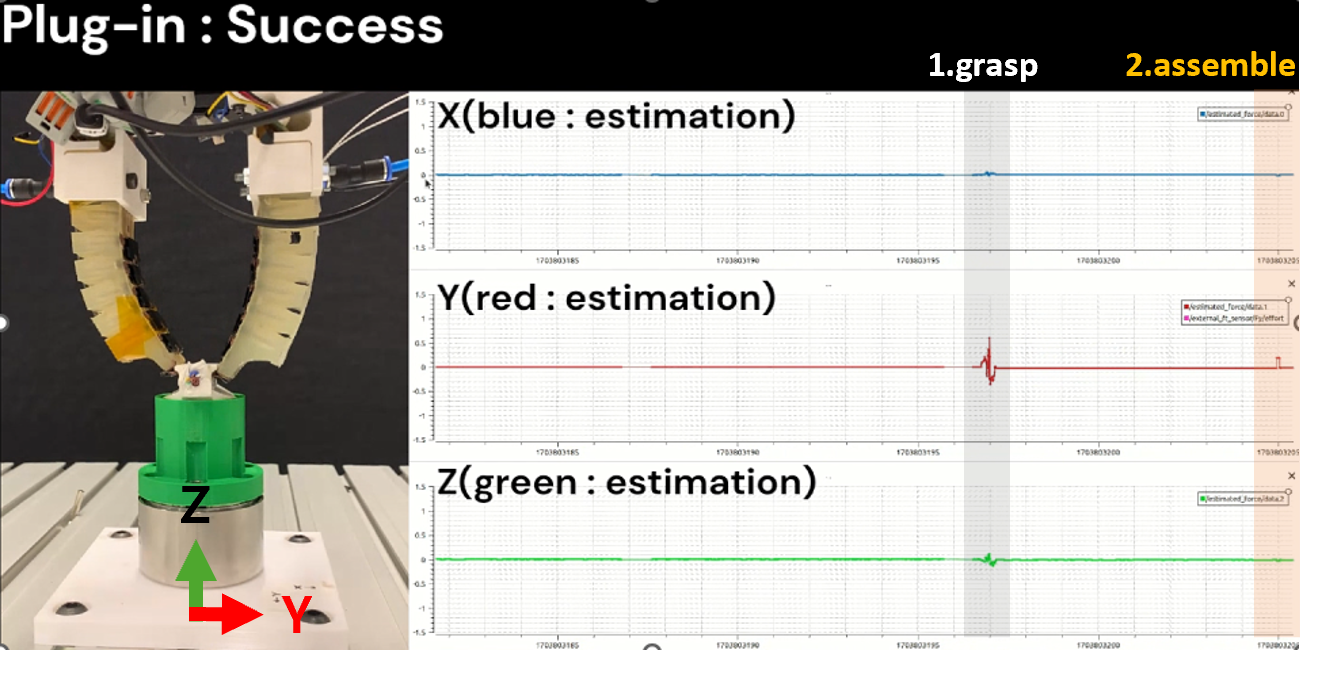} \\
    \includegraphics[width=0.8\columnwidth]{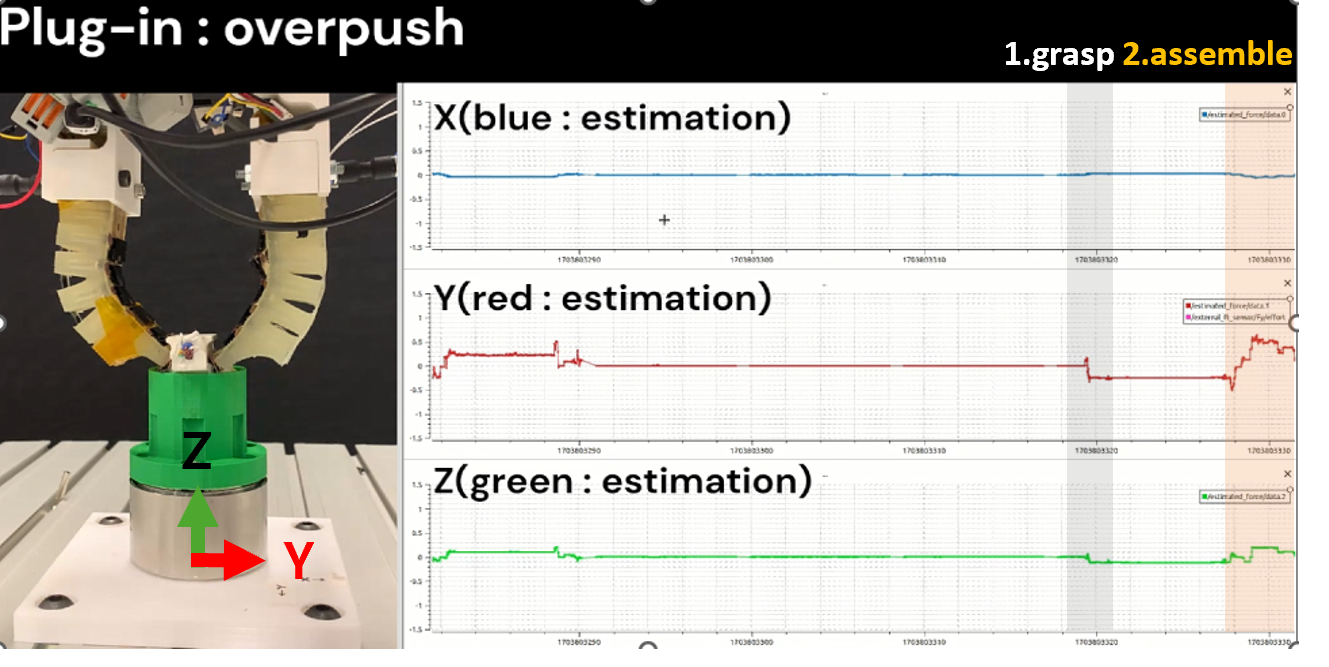} 
    \includegraphics[width=0.8\columnwidth]{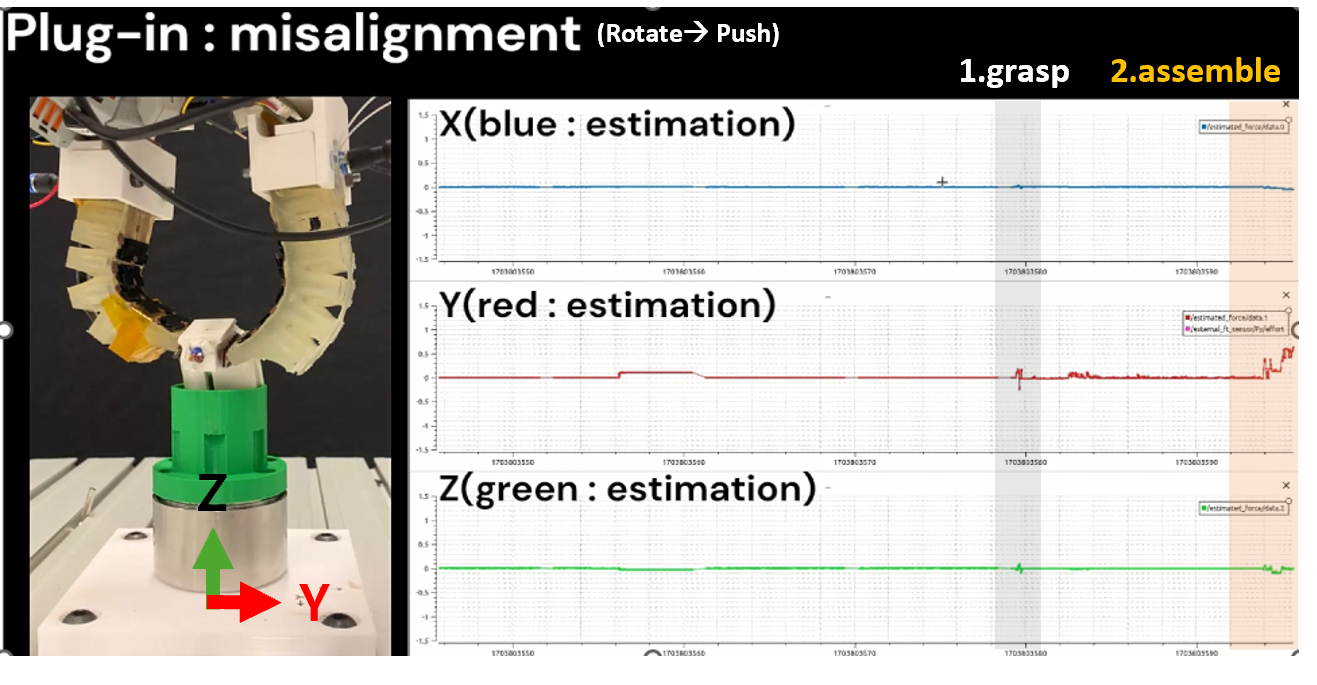} 
    \caption{Plug insertion task with estimated forces on the right. The forces during grasp actuation are detected in all cases, while the assembly forces are visible in the larger deformations of (middle) and (bottom). Please see the attached video for the online force measurements.}
\label{fig:validate_assm}
\end{figure}

In the second application, the task involves electrical plug insertion. A diverse array of manipulation outcomes, including successful plug-in, grasping failure due to over-pushing, and misalignment, are tested as seen in Figure \ref{fig:validate_assm} and in the attached video\footnote{https://www.youtube.com/watch?v=XSTeJP6nMxE}. After grasping the plug, the robot moves up, and for misalignment rotates about the Z-axis. Grasping success and failure instances are readily discerned through variations in force detection, with higher force readings indicative of failure occurrences.

In all cases, the changes grasp event can be distinguished in the force estimates. In the overpush (middle) and misalignment (lower) cases, the larger contact forces can be clearly distinguished in the final phase of the plots. The lower forces involved in the perfect success case (top) cannot be seen, due to the lower sensitivity of the sensors and force estimation model.

% 3x grasping -> detecting the changes in the manipulation task

\section{Conclusion}[t]
This paper presented a machine learning approach to estimating the forces applied by soft fingers. We validated the effectiveness of normal force in predicting four contact states—non-contact, contact, slip, and stick—and found it robust to changes in object type and robot pose. A model for estimating the total force was then proposed and validated on various objects. While the model demonstrated some generalization to new objects, there was a slight loss in accuracy. Additionally, the model output was found to change with internal and external contact modes in slip and assembly finger applications.

The approach has limitations that contribute to inaccuracies. The pressure sensors do not measure shear forces, and significant finger deformation is required to estimate these forces from the strain sensors. To mitigate this issue, using robot position information as an input feature could provide crucial data about the direction of deformation. Furthermore, our training data was recorded in static scenarios, spanning 3,738 seconds. Including dynamic scenarios in the training set could help address this limitation.

The interaction between fingers during grasping also poses challenges. While normal force magnitude is mitigated, frictional forces are amplified, complicating slip detection. Training each finger individually could improve the accuracy of contact state estimation.

In conclusion, while the proposed method for discriminating contact states is promising, enhancing the training data and adjusting features are crucial for improving slip detection. Future work should focus on refining the model to better capture complex interactions and explore using these estimates for grasp improvement and task monitoring in manipulation tasks.

\bibliographystyle{IEEEtran}
\bibliography{lib_kevin,lib_valentyn, lib_hun}

\end{document}